\documentclass[times,twocolumn,final,authoryear]{elsarticle}
\usepackage{prletters}
\usepackage{graphicx}
\usepackage{microtype}
\usepackage{subfigure}
\usepackage{float}
\usepackage{tabulary,tabularx, xcolor}
\usepackage{amsfonts,amsmath,amssymb}
\usepackage{algorithm, algpseudocode}
\usepackage{rotating}
\usepackage{multirow}
\usepackage{verbatim}
\usepackage{array}
\usepackage{lineno,hyperref}
\usepackage{url}
\usepackage{xcolor}
\usepackage[markup=underlined]{changes}

\definecolor{newcolor}{rgb}{.8,.349,.1}

\journal{Pattern Recognition Letters}

\begin{document}

\clearpage
\thispagestyle{empty}
\ifpreprint
  \vspace*{-1pc}
\fi

\begin{frontmatter}
\title{Block based Singular Value Decomposition approach to matrix factorization for recommender systems}
\author[venk]{Prasad Bhavana\corref{cor1}}
\ead{17mcpc14@uohyd.ac.in, prasadbgv@gmail.com}
\author[raj]{Vikas Kumar}
\ead{vikas007bca@gmail.com}
\author[venk]{Vineet Padmanabhan}
\ead{vineetcs@uohyd.ernet.in}
\cortext[cor1]{Corresponding author:
Tel.: +91-40-23014559;  
fax: +91-40-23134012;}
\address[venk]{Artificial Intelligence Lab, School of Computer and Information Sciences, University of Hyderabad, Hyderbad-500046, AndhraPradesh, India}
\address[raj]{Central University of Rajasthan, Rajasthan, India}

\begin{abstract}

With the abundance of data in recent years, interesting challenges are posed in the area of recommender systems.  Producing high quality recommendations with scalability and performance is the need of the hour. Singular Value Decomposition(SVD) based recommendation algorithms have been leveraged to produce better results. In this paper, we extend the SVD technique further for scalability and performance in the context of 1) multi-threading 2) multiple computational units (with the use of Graphical Processing Units) and 3) distributed computation. We propose block based matrix factorization (BMF) paired with SVD. This enabled us to take advantage of SVD over basic matrix factorization(MF) while taking advantage of parallelism and scalability through BMF. We used Compute Unified Device Architecture (CUDA) platform and related hardware for leveraging Graphical Processing Unit (GPU) along with block based SVD to demonstrate the advantages in terms of performance and memory. 

\end{abstract}


\end{frontmatter}

\section{Introduction}


Recommending items to a user based on his/her past preferences is a well-studied problem in the area of Machine Learning (ML). 
In the recent past, several techniques have been proposed to address the problem of recommendation. These techniques are primarily grouped into three major categories namely content-based recommendation, collaborative recommendation (filtering) and hybrid recommendation. In the content-based approach, the recommendation is made by using the profile information of a user and an item. For example, in movie recommendation the movie profile can contain the \textit{genre} information and then based on the user interest for the \textit{genre}, a particular movie is recommended. In collaborative filtering, an item is recommended to a user based on his/her past preferences and the preference information of other similar users. For example, in movie recommendation the rating information can be used to find other similar users. The hybrid approach can be seen as a combination of both the content-based and collaborative-based model.

The content-based collaborative filtering has certain limitations and cannot be applied to situations where the item features are not meaningful or to situations where there is a need to capture the change in user interests over time. Collaborative filtering alleviates the above mentioned challenges as it only requires the preference (implicit or explicit) information for recommendation. There are several approaches of collaborative filtering which can be further grouped into memory-based and model-based collaborative filtering. In memory-based collaborative filtering, the recommendation is made by finding the similarity score between a user and items. Based on the similarity score, a list of top-$K$ items are recommended to a user. Recommending items based on nearest neighborhood is a typical example of memory-based collaborative filtering. At first, given a target user, a set of $k$-similar users are first identified based on the observed preferences. Then the model recommends a set of items based on the likes of similar users. In most of the real-world data sets, the observed preferences are very sparse and there are very few items rated by a set of common users. This leads to the calculation of unreliable similarity values and in such scenarios memory-based models perform very poorly~\cite{su2009survey}. On the other hand, in model-based collaborative filtering, the goal is to learn the latent (hidden) representation of the users and the items. Based on the affinity in the latent space representation, an item is either recommended or not recommended to a user. 

Model-based collaborative filtering can be visualized as a matrix completion task. Given a data set of $m \times n$ user-item ratings with $n$ number of users and $m$ number of items, the aim of collaborative filtering is to predict unobserved preference of users for items~\cite{KUMAR20171,KUMAR201762}. Matrix factorization (MF) is one of the prominent techniques for matrix completion. The objective of matrix factorization is to learn latent factors $U$ (for users) and $V$ (for items) simultaneously. The latent factors are used for approximation of the observed entries, so as to evaluate the model, using some loss measure. The latent factors thus derived are used further to predict the unobserved entries.
With each passing year, more and more preference data gets generated and the task of recommendation becomes more challenging. 
With the exponential increase in the preference data, the major challenge is to provide a more accurate recommendations with less computational effort. Several approaches have been proposed in the literature that take advantage of availability of high volume of data for better and accurate recommendation. Though there are a few important proposals, research on scalability, parallelism and distributed computation to handle large volumes of data has not attracted much attention of researchers. 
For instance, Mackey et al.~\cite{NIPS20114486} proposed a divide and conquer based approach for parallelism in matrix factorization by treating factorization of each sub-matrix as a sub-problem and thereafter combining the results. This approach resulted in noisy factorization. Similarly, 
in~\cite{Zhang:2013:LMF:2488388.2488520} a localized factorization is proposed for recommendation on a block diagonal matrix. In \cite{Du2017}, a divide and conquer strategy based Non-Negative Matrix Factorization (NNMF) is proposed for fast clustering and topic modeling. To make the model scalable from rank-2 to rank-$k$, the authors proposed to use a binary tree structure of the data items. 
In \cite{7727641}, a block kernel based approach of matrix factorization is proposed for the face recognition task. Nikulin et al.~\cite{NIKULIN2011773} proposed a fast gradient based  matrix factorization algorithm for use in statistical analysis. From what has been said till now it can be noted that matrix factorization based approach is a popular strategy for recommendation and is still an active area of research. 
There are a few other notable proposals that handle large data sets either addressing parallelism or distributed computation but not both~\cite{DBLP:journals/corr/SchelterSZ14, DBLP:conf/kdd/GemullaNHS11, Yun:2014:NNS:2732967.2732973, DBLP:conf/edbt/LiTS13, Zhuang:2013:FPS:2507157.2507164, DBLP:conf/nips/RechtRWN11, DBLP:conf/kdd/OhHYJ15, DBLP:conf/icdm/YuHSD12, DBLP:conf/pakdd/ChinZJL15}. 
In this paper, we propose a variant of Singular value Decomposition (SVD) called \textit{Block} based Singular Value Decomposition for large scale recommendation task. We also demonstrate how \textit{parallelism} can be achieved by employing Graphical Processing Unit (GPU).

The rest of the paper is organized as follows. Section $2$ summarizes the well-known existing Singular Value Decomposition approach. In Section $3$, we briefly discuss the Block based Matrix Factorization approach and how parallelism can be achieved through the GPU. We introduce the proposed Block based variant of SVD (BSVD) in Section $4$. The advantage of the proposed approach over the  existing method is reported in Section $5$. Finally, Section $6$ concludes and indicates several issues for future work.

\section{Singular Value Decomposition}

Singular value decomposition (SVD) is closely related to a number of mathematical and statistical techniques that are used in a wide variety of fields, including eigen vector decomposition, spectral analysis, factor analysis, etc. SVD is applied to a large variety of applications including dimensionality reduction~\cite{doi:10.1002/(SICI)1097-4571(199009)41:6<391::AID-ASI1>3.0.CO;2-9, Berry95usinglinear}, computer vision~\cite{1613062}, signal processing \cite{1613062}, \cite{DELATHAUWER200431} etc. One of the important applications of SVD is a matrix completion problem wherein given a data matrix $X \in \mathbb{R}^{m~\times~n}$ with $m$ rows and $n$ columns, the goal is to derive a set of uncorrelated low-dimensional factors in the  ``\textit{eigen rows}" $\times$ ``\textit{eigen columns}" space. The numerical rank is much smaller than $m$ and $n$, and hence, factorization allows the matrix to be stored inexpensively. The original data matrix then can be recovered with these low-dimensional factor matrices 

Given a $m \times n$ size matrix $X$, the $SVD(X)$ is defined as.
\begin{equation}
\label{SVDEQ}
    SVD(X) = USV^T
\end{equation}
where $U$, $S$ and $V$ are  of  dimensions $m \times m, m \times n$, and $n \times n$, respectively. 
The matrices $U$ and $V$ are orthogonal matrices and $S$ is a diagonal matrix, called the singular matrix. The diagonal entries ($s_1,s_2,...,s_n$) of $S$ are in incremental order; i.e., $s_1 \geq s_2 \geq ... \geq s_m > 0$. These matrices $U$, $S$, and $V$ represent a breakdown of the original relationships into linearly independent components or factors. In the diagonal matrix $S$, many of the entries are very small, and may be ignored, leading to an approximate model that contains many fewer dimensions. With $k$ number of non-zero entries (the size of reduced dimensional space or most significant values), the effective  dimensions  of these three matrices $U$, $S$, and $V$ are $m \times k$, $k \times k$, and $n \times k$, respectively. 
We can choose a small rank (k) and extract a matrix of exactly that rank from the SVD. The resulting matrix will still approximate the original matrix. Therefore decreasing the rank will just smooth out the entries in the recovered matrix by forcing them to be linear combinations of only a few basis vectors and at the same time match our sparsely observed ratings as closely as possible. The result can be represented geometrically by a spatial configuration in which the dot product or cosine between user and item vectors represent estimated similarity of the two objects.

\subsection{SVD for recommendations}
Recommender system is one of the prominent applications of SVD where the aim is to find and fit a useful model of the relationship between users and items. The idea is to learn the underlying parameters of the model including the latent factors of users and items using the observed rating preferences. 
Using the learnt latent factors, we predict the association between users and items for which the preferences were unobserved. 
However, computing SVD of a user-item matrix is expensive and requires a large amount of memory and computational effort. For reasonable number of users and/or items, it may not even be possible to fit the matrix in memory to begin with. In order to compute the SVD efficiently, in \cite{Sarwar02incrementalsingular} a practical approach to leveraging incremental computation of SVD for recommender systems is proposed. The paper proposes \emph{folding-in} based SVD technique for factorization. Koren et al.~\cite{Koren091the} extended the incremental computation of SVD at an element level to capture the  temporal changes in user and item biases. 

As given in Eq.~(\ref{SVDEQ}), the goal of SVD computation is to learn the factor matrices $U$, $S$, and $V$. For the sake of simplicity and meaningful explanation, we could consider matrix $S$ as an identity matrix. It is a diagonal matrix, so it simply acts as a scalar on $U$ or $V^T$. Hence, we can assume that we have merged the scalar factors into both the matrices $U$ and  $V$ during the approximation. So the matrix factorization simply becomes $X=U \times V^T$. Considering the rating value $x_{u i}$ as the result of a dot product between two vectors: a vector $p_u$ which is a row of $U$ and is specific to the user $u$, and a vector $q_i$ which is a column of $V^T$ and is specific to the item $i$ : $x_{u i}=p_u \times q_i$. So, the SVD of $X$, is merely modeling the rating of user $u$ for item $i$ as

\begin{equation}
x_{u i} = \sum_{f \in latent factors} ( \textrm{affinity of u for f} \times \textrm{affinity of i for f})
\label{eq:1}
\end{equation}

In other words, if $u$ has a taste for factors that are endorsed by $i$, then the rating $r_{ui}$ will be high. However, due to the elimination of $S$, the typical/general user, item biases represented by the singular values are eliminated from prediction of unknown ratings. 
This causes deviation in the Root Mean Square Error(RMSE) computation for unknown ratings. An alternative way to \emph{factor-in} bias is proposed in \cite{Koren091the}. The authors proposed adding them back into the equation as a linear combination, which is represented as
\begin{equation}
x_{u i} = p_u \times q_i^T + bu_u + bi_i
\label{eq:3}
\end{equation}
where $bu_u$ represents a singular value of user bias for $u$ and similarly $bi_i$ represents a singular value of item bias for $i$. 

\subsection{Stochastic Gradient Descent (SGD) approach to SVD}

In most of the real-world applications, the rating matrix $X$ is partially observed and for such matrices the computation of $XX^T$ and $X^TX$ do not exist, so their eigenvectors do not exist either. 
Hence, the SVD computation is not defined. In such situations, the latent factor matrices $U$ and $V$ can actually be learnt if we can find all the vectors $p_u$, $q_i$, $bu_u$ and $bi_i$ such that the $p_u$ make up the rows of $U$ and the $q_i$ make up the columns of $V^T$. The related optimization problem can be represented as.

\begin{equation}
\min_{p_u, q_i, bu_u, bi_i} J =  \sum_{{u i} \in \Omega} (x_{u i} {-} p_u . q_i^T {-} bu_u {-} bi_i)^2
\label{eq:4}
\end{equation}
where $\Omega$ is set of observed entries. However, this optimization problem is not convex and hence requires an approximation technique to arrive at a solution. SGD (Stochastic Gradient Descent) is one of the techniques that can find the approximate solution. The optimal values of the latent variables can be obtained by minimizing Eq.~\ref{eq:4}.
The gradients with respect to $p_u$ and $q_i$ (in vector notation) are given by
\begin{equation}
    \frac{\partial J}{\partial p_u} = {-2}q_i (x_{u i} {-} p_u . q_i^T {-} bu_u {-} bi_i)
\label{eq:5}
\end{equation}
\begin{equation}
    \frac{\partial J}{\partial q_i} = {-2}p_u (x_{u i} {-} p_u . q_i^T {-} bu_u {-} bi_i).
\label{eq:6}
\end{equation}
Similarly, the gradients with respect to $bu_u$ and $bi_i$ (in vector notation) are given by
\begin{equation}
    \frac{\partial J}{\partial bu_u} = {-2}(x_{u i} {-} p_u . q_i^T {-} bu_u {-} bi_i)
    \label{eq:7}
\end{equation}

\begin{equation}
    \frac{\partial J}{\partial bi_i}  = {-2}(x_{u i} {-} p_u . q_i^T {-} bu_u {-} bi_i)
    \label{eq:8}
\end{equation}

When matrix completion problem is viewed as supervised learning with $\Omega$ as the training set, it becomes necessary to ensure that overfitting is avoided. This can be done by minimizing the regularized loss function and thereby having the following formulation.

\begin{multline}
\min_{p_u, q_i, bu_u, bi_i} J =  \sum_{{u i} \in \Omega} (x_{u i} {-} p_u . q_i^T {-} bu_u {-} bi_i)^2 \\
+  \beta_1 \sum_{{u} \in \Omega} {p_u}^2 + \beta_2 \sum_{{u} \in \Omega} {q_i}^2 + \beta_3 \sum_{{u} \in \Omega} {bu_u}^2 + \beta_4 \sum_{{i} \in \Omega} {bi_i}^2 
\label{eq:9}
\end{multline}

\noindent With the inclusion of regularization parameters, the update equations for \ref{eq:5}, \ref{eq:6}, \ref{eq:7} and \ref{eq:8} can be rewritten with learning co-efficient $\alpha_1$, $\alpha_2$, $\alpha_3$, $\alpha_4$ and regularization coefficients $\beta_1$, $\beta_2$, $\beta_3$, $\beta_4$ as shown below:

\begin{equation}
    p_u \leftarrow p_u + \alpha_1 . q_i (x_{u i} {-} p_u . q_i^T {-} bu_u {-} bi_i - \beta_1 p_u) 
    \label{eq:10}
\end{equation}

\begin{equation}
    q_u \leftarrow q_u + \alpha_2 . p_u (x_{u i} {-} p_u . q_i^T {-} bu_u {-} bi_i - \beta_2 q_u)
    \label{eq:11}
\end{equation}

\begin{equation}
    bu_u \leftarrow bu_u + \alpha_3 (x_{u i} {-} p_u . q_i^T {-} bu_u {-} bi_i - \beta_3 bu_u)
    \label{eq:12}
\end{equation}

\begin{equation}
    bi_i \leftarrow bi_i + \alpha_4 (x_{u i} {-} p_u . q_i^T {-} bu_u {-} bi_i - \beta_4 bi_i)
    \label{eq:13}
\end{equation}


\section{Block based approach to Matrix Factorization}\label{ss:bmf}

Consider $X \in \mathbb{R}^{m~\times~n}$ be a rating matrix with ratings for $m$ users and $n$ items. The matrix factorization (MF) approach is visualized as an estimation of the data matrix $X \approx UV^T$ where latent factor matrices $U \in \mathbb{R}^{m~\times~k}$ and $V \in \mathbb{R}^{n~\times~k}$ (for some chosen dimension $k$) are derived from the given data. The given data matrix can be represented in block notation as given in~\eqref{eq:datax}. The representation is based on matrix formation with blocks of equal dimension and if required, zeros can be padded to the data matrix to ensure all the blocks are of equal size.

\begin{equation}
X = \begin{bmatrix}
	X_{11} & X_{12} & \dots & X_{1j} & \dots  & X_{1J} \\
	\vdots & \vdots & \dots & \vdots & \ddots & \vdots \\
	X_{i1} & X_{i2} & \dots & X_{ij} & \dots  & X_{iJ} \\
	\vdots & \vdots & \dots & \vdots & \ddots & \vdots \\
	X_{I1} & X_{I2} & \dots & X_{Ij} & \dots  & X_{IJ}
\end{bmatrix}\label{eq:datax}
\end{equation}

\begin{figure}[ht!]
\begin{center}
  \includegraphics[width=\linewidth,height=4.5in]{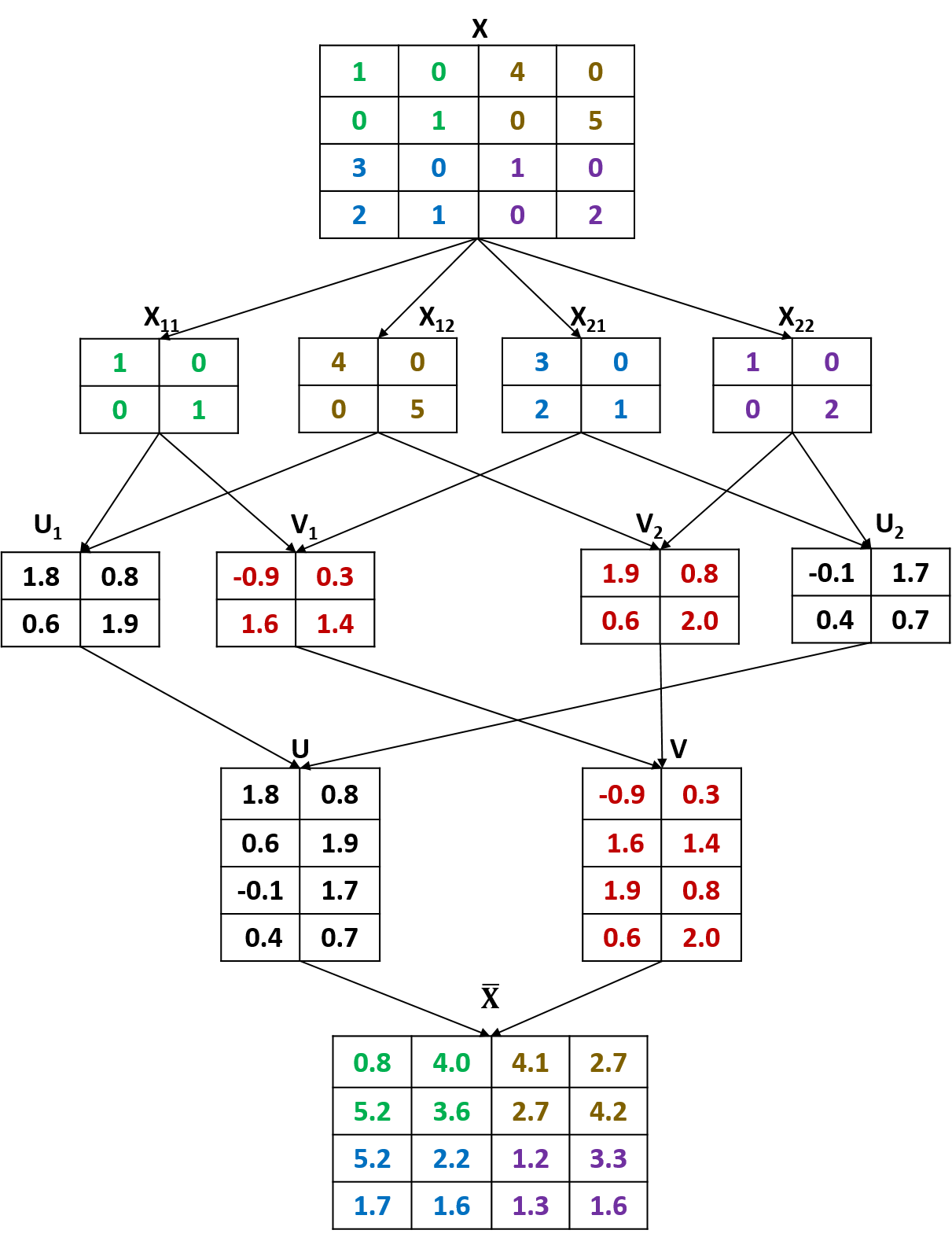}
  \caption{Example of Block Matrix Factorization}
  \label{fig:example}
  \end{center}
\end{figure}

The Block based approach to Matrix Factorization (BMF) considers each block as an individual matrix. It then factorizes the block for one iteration and takes the latent features of each of these blocks as a starting point for approximation of the latent features for the relevant blocks there after. Figure~\ref{fig:example} demonstrates a simple example wherein $X$ is divided into $4$ blocks and each of these blocks are factorized individually so as to combine together to form $U$ and $V$ that exactly explain $X$.

 BMF considers each element exactly once per iteration, with the difference in change of sequence of processing of elements. As MF does not constrain the sequence in which the data elements are processed, the convergence of BMF is expected to be equivalent to MF.  
 As a limiting condition, matrix factorisation can be viewed as BMF where each element is a different block or where the entire matrix is considered as a single block. 
 

\begin{figure}[ht!]
\begin{center}
  \includegraphics[width=3.2in, height=2.8in]{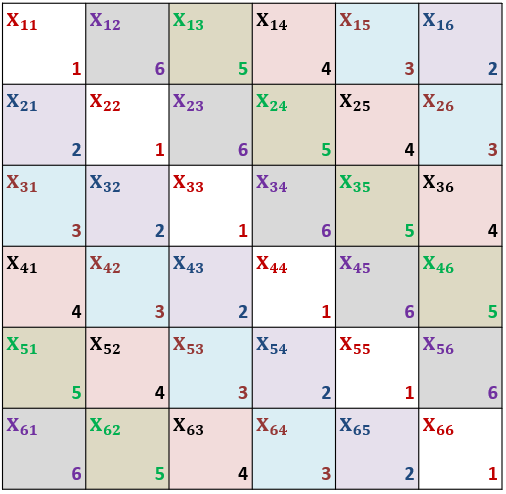}
  \caption{An example scenario of parallel BMF}
  \label{fig:bmf}
  \end{center}
\end{figure}
\vspace{6mm}
\noindent \textbf{Parallelism: }
Using the matrix blocks, $U_i$ and $V_j$ matrices can be derived simultaneously for multiple blocks. This is achieved by first identifying the data blocks whose latent factors do not depend on each other and dedicating a computation unit for each such block. Figure~\ref{fig:bmf} shows one such example with a $6\times6$ block matrix wherein blocks represented with the same number (placed at lower right corner of the cell) can be simultaneously factorized; i.e., blocks on the diagonal are processed in the first parallel step and then the factorization is moved on to the blocks below them while considering the entire column as a loop starting from the main diagonal block.\\

\noindent \textbf{Parallelism through GPU computation: }
In order to fully leverage parallelism, the multi-threading capabilities of the GPU can be utilized. In~\cite{Tan2018MatrixFO}, a GPU accelerated matrix factorization is proposed for 
the approximate Alternative Least Square (ALS) algorithm. The authors propose to use SGD for the optimization. GPU accelerated Non-Negative Matrix Factorization (NNMF) for Compute Unified Device Architecture (CUDA) capable hardware has been proposed in~\cite{koitka2016nmfgpu4r}. Similarly, in ~\cite{10.1007/978-3-642-31178-9_15}, NNMF with GPU acceleration is used for text mining purposes. From the literature it can be found that various matrix factorization based approaches have been proposed which includes parallel as well distributed frameworks for scaling up the factorization process as mentioned in the Introduction section.
Our approach is to make use of block based approach for parallelism and combine it with GPU computation to demonstrate the advantages of the combined approach.

\section{Block based approach to SVD}

\begin{algorithm}[ht!]
\footnotesize
\begin{algorithmic}[1]
\Require Input: Data matrix $X \in \mathbb{R}^{\{m \times n\} }$, number of features $k$
\Ensure Initialize: latent feature matrices with random values $U \in \mathbb{R}^{\{m \times k\}}, V \in \mathbb{R}^{\{n \times k\} }$
\State Let $I, J$ be two constants such that $X$ is represented by $I \times J$ number of sub-matrices
\State Represent data matrix $X$ as block matrix with sub-matrices $X_{i j}$ where $i \in 1..I$ and $j \in 1..J$. Similarly, represent feature matrices $U, V$ as block matrices with sub-matrices $U_i, V_j$ where $i \in 1..I$ and $j \in 1..J$. 
\State Let $bu \in \mathbb{R}^{\{1 \times m\}}, bi \in \mathbb{R}^{\{1 \times n\}}$ be two vectors to represents biases of uses, items respectively
\State Let STEPS be a constant representing maximum iterations for factorization of SVD and $\alpha_1, \alpha_2, \alpha_3, \alpha_4$ be the learning rates, $\beta_1, \beta_2, \beta_3, \beta_4$ be the regularization factors and $\delta$ the minimum deviation of error between iterations

\For{step 1 to STEPS}

	\For{each block sub-matrix ${X_{\{i \times j\}}}$ }

		\State{${U_i}, {V_j}, bu, bi \gets BLOCK\_SVD(X_{i j}, {U_i}, {V_j}, bu, bi$, $k$, $\alpha_1$, $\alpha_2$, $\alpha_3$, $\alpha_4$, $\beta_1$, $\beta_2$, $\beta_3$, $\beta_4$)}

	\EndFor
	\State{Terminate if RMSE improvement is $< \delta$}

\EndFor
\State{Return latent feature matrices $U, V $}
\caption{Block based approach to factorization of SVD}
\label{alg:one}
\end{algorithmic}
\end{algorithm}

\begin{algorithm}
\footnotesize
\begin{algorithmic}[2]
\Function{block\_svd}{$X_{ij}, {U_i}, {V_j}, bu, bi, k, \alpha_1, \alpha_2, \alpha_3, \alpha_4, \beta_1, \beta_2, \beta_3, \beta_4$}
\For{each row $r$ in block $X_{ij}$}
\For{each column $c$ in block $X_{ij}$}
    \If {$X_{ij}[r,c] > 0 $}
    \State{$err \gets X_{ij}[r,c] - bu[r] - bi[c] - {U_i}[r,*] . {V_j}[c,*]^T $}
    \State{$bu[r] \gets bu[r] + \alpha_1$ (err - $\beta_1 . bi[c]$)}
    \State{$bi[c] \gets bi[c] + \alpha_2$ (err - $\beta_2 . bu[r]$)}
    \For{each latent factor $k$}
         \State{${U_i}[r,k],\gets {U_i}[r, k] + \alpha_3 (err . {V_j}[c,k] - \beta_3 . {U_i}[r, k]$)}
         \State{${V_j}[c,k],\gets {V_j}[c,k] + \alpha_4 (err . {U_i}[r,k] - \beta_4 {V_j}[c,k]$)}
    \EndFor
    \EndIf
\EndFor
\EndFor
\State{Return latent feature block matrices, biases ${U_i}, {V_j}, bu, bi $}
\EndFunction
\caption{SVD based matrix factorization for a block}
\label{alg:svd}
\end{algorithmic}
\end{algorithm}

The fundamental idea behind block based SVD approach is to employ block based factorization with an SVD kernel. A broad outline of the approach is given in Algorithm~\ref{alg:one}. The $BLOCK\_SVD()$ function in the Algorithm \ref{alg:svd} encapsulates implementation of SVD based matrix factorization at a block level for one iteration only. Considering the number of steps and $k$ as constants and as the two for-loops (iterating STEPS number of times, and iterating for each sub-matrix) already contribute towards number of ratings, hence, the time complexity of the algorithm remains as $O(n)$ for $n$ ratings. Considering the block size as constant, the space complexity is $\approx ( m\times n + m\times k + m\times k) \times c$ $\approx O(c)$ (constant)

The implementation of the Block based SVD algorithm is made available for public (open source) on Github~\footnote{https://github.com/17mcpc14/blockgmf}.

\section{Experiments}\label{sssec:analysis}
In this section, we discuss the experimental setup and report the related results. The experiments mentioned below are conducted on a shared hardware with Intel(R) Xeon(R) CPU E5-2640 v3 @ 2.60GHz and with Nvidia Tesla M40 GPU with 24GB of CPU memory and 8GB of GPU memory. The programming environment is Python 2.7.5 that leverages CUDA with driver version 8.0 and with PyCUDA application programming interface version 1.8.
To demonstrate the scalability of the proposed approach, we carried out experiments on five publicly available real-world data sets namely \textit{MovieLens - 100K}\footnote{\url{https://grouplens.org/datasets/movielens/} \label{movielens}}, \textit{MovieLens - 1M}\ref{movielens}, \textit{MovieLens - 10M}\ref{movielens}, \textit{MovieLens - 20M}\ref{movielens} and \textit{Jester}\footnote{\url{https://www.ieor.berkeley.edu/~goldberg/jester-data/}}. The data sets are randomly divided into training- and test-set. We consider $80\%$ of the observed entries to train the model and remaining $20\%$ to test the performance. The experiments are repeated $3$ times and the mean and standard deviation of the results have been presented. We compare our proposed method with two well-known algorithms, Probabilistic Matrix Factorization (PMF)~\cite{Salakhutdinov:2007:PMF:2981562.2981720} and SVD. For fair comparison, we further adopted Stochastic Gradient Descent search to optimize the objective function of PMF and SVD.

\noindent \textbf{Hyper-parameters: }For all the experiments reported in the subsections that follow, the learning rate ($\alpha$) and regularization parameter ($\beta$) in PMF are fixed to $0.0001$ and $0.01$, respectively. In SVD, the parameters $\alpha$ and $\beta$ for the user latent feature computation are set to $0.019$, $0.019$, respectively. Similarly. for the item latent feature computation $\alpha$ and $\beta$ are kept fixed to $0 .004$ and $0.019$, respectively. For user biases computation $\alpha=0 .004$, $\beta=0.019$ are used. For item biases computation $\alpha=0.013$, $\beta=0.007$ are used. The minimum difference of error between steps ($\delta$) considered for early termination for all the algorithms is taken as $0.0001$. The number of latent factors is kept fixed to $13$. 
\begin{figure}[ht!]
  \includegraphics[width=\linewidth, height = 2.7in]{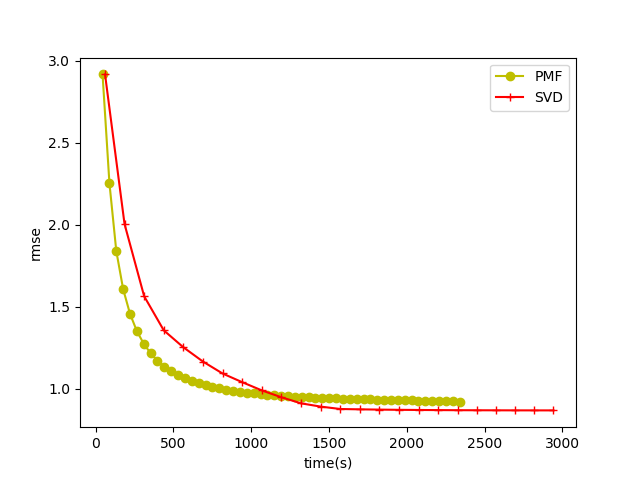}
  \caption{PMF vs SVD for MovieLens 1M data set}
  \label{fig:pmfvssvd}
\end{figure}
\subsection{Experimental results}

Figure \ref{fig:pmfvssvd} shows the comparative performance advantage of SVD over basic PMF. It can be seen that the RMSE convergence of SVD is better than PMF. We would like to mention that the convergence time may not be an indicator of performance, as the learning rates and the number of parameters are different for each algorithm and are defined as per the requirements of each algorithm. The time per iteration and RMSE values of the algorithms are detailed in Table \ref{tab:standard}. \\

\begin{figure}[!t]
\centering
\begin{minipage}[b]{0.49\textwidth}
  \includegraphics[width=\linewidth, height = 2.7in]{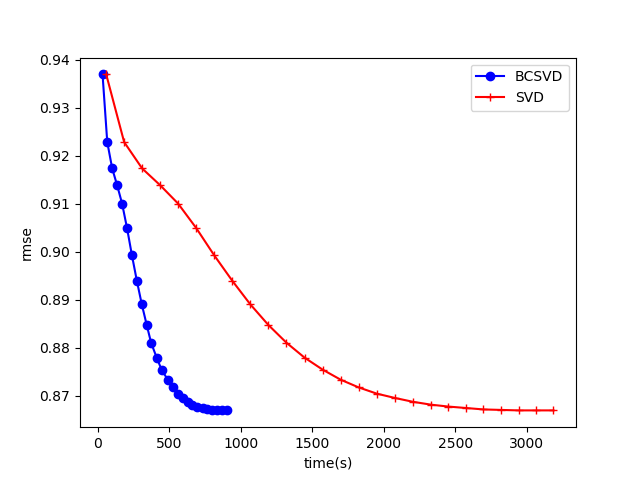}
  \caption{SVD vs Block based SVD for MovieLens 1M data set}
  \label{fig:svdvsbsvd}
\end{minipage}
\hfill
\begin{minipage}[b]{0.49\textwidth}
  \includegraphics[width=\linewidth, height = 2.7in]{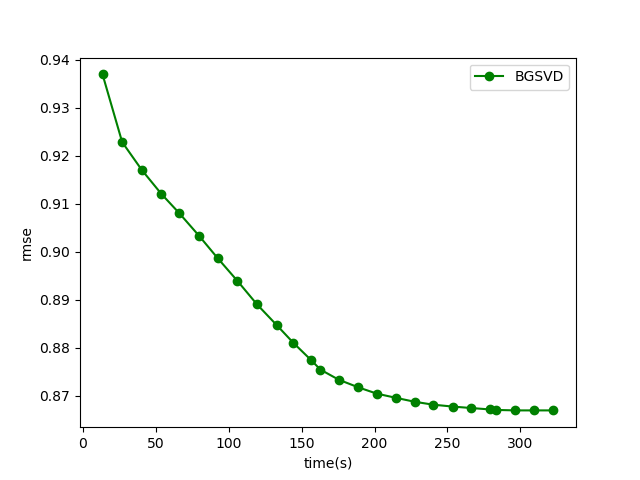}
  \caption{BGSVD run on MovieLens 1M data set}
  \label{fig:bsvdvsbgsvd}
\end{minipage}
\end{figure}

\noindent \textbf{SVD vs Block based SVD: } As discussed in Section \ref{ss:bmf}, applying block based approach on any MF enables us to process the data in parallel without compromising on the outcome. We have implemented Block based CPU variant of SVD (BCSVD) in line with the update equations discussed in Section \ref{sssec:svd} and with the use of block based approach. In order to achieve parallelism, we have implemented multiple CPU threads to work in tandem. Each thread is dedicated to processing one block at a time, while splitting the data matrix into square block matrices. Combinations of different number of blocks have been experimented. While increasing the number of blocks increases parallelism, however, we observed a performance drop beyond a critical point. We hypothesize that this observation is due to increased pagination as the threads compete for resources. As our objective is to implement and analyze BMF, we limited the scope to parallelism through BMF. Hence, we configured just 8 parallel threads in accordance with the number of CPU cores available on our hardware. Accordingly we split the data matrix into $8\times 8$ block matrix. Figure \ref{fig:svdvsbsvd} demonstrates the comparative analysis of SVD with BCSVD. We used MovieLens 1 million data set with 6040 users and 3900 movies, with each block of $755 \times 486$ dimension. We used 64 bit float data types for $U, V$ matrices and 32 bit integer data type for ratings. For MovieLens 1M data set, the maximum memory required for each block including the rating data and latent feature vectors, is estimated to be of \~1.513 Mega Bytes (MB). It is observed that the memory needed per block is much lower (\~0.3 MB per block) due to the sparse nature of the ratings. The memory needed for the algorithm is estimated to be constant and is a maximum of \~12.1 MB. This estimate excludes the memory needed for the code, program variables etc. Table~\ref{tab:standard} lists the comparative analysis of time taken per iteration and test RMSE for different variants of MF listed above. The results confirm our hypothesis that the block based approach has no impact on the quality of outcome but only on the performance.\\

\begin{table*}[ht!]
\centering
\begin{tabular}{c|c|r|r|r|r|r|}
\cline{2-7}
\multicolumn{1}{l|}{}                                                                                           & \multicolumn{1}{l|}{Data set --\textgreater{}} & \multicolumn{1}{c|}{\begin{tabular}[c]{@{}c@{}}MovieLens\\  100K\end{tabular}} & \multicolumn{1}{c|}{\begin{tabular}[c]{@{}c@{}}MovieLens \\ 1M\end{tabular}} & \multicolumn{1}{c|}{\begin{tabular}[c]{@{}c@{}}Jester \\ 4M\end{tabular}} &
\multicolumn{1}{c|}{\begin{tabular}[c]{@{}c@{}}MovieLens \\ 10M\end{tabular}} & \multicolumn{1}{c|}{\begin{tabular}[c]{@{}c@{}}MovieLens \\ 20M\end{tabular}}\\ 
\hline
\multicolumn{1}{|c|}{\multirow{4}{*}{\begin{tabular}[c]{@{}c@{}}Training Time \\ (sec/iteration)\end{tabular}}} & PMF & 4.80 $\pm$ 0.61 & 44.14 $\pm$ 0.42 & 126.98 $\pm$ 1.34 & 490.79 $\pm$ 3.36 & 967.84 $\pm$ 5.41
\\ \cline{2-7} 
\multicolumn{1}{|c|}{} & SVD & 5.24 $\pm$ 0.68& 49.56 $\pm$ 0.21 & 139.08 $\pm$ 1.20 & 510.16 $\pm$ 4.16 & 1173.18 $\pm$ 5.9
\\ \cline{2-7} 
\multicolumn{1}{|c|}{} & BCSVD & 3.53 $\pm$ 1.01 & 34.53 $\pm$ 2.32 & 98.124  $\pm$ 3.54 & 193.69  $\pm$ 5.63 & 682.14  $\pm$ 6.98
\\ \cline{2-7} 
\multicolumn{1}{|c|}{} & BGSVD & \textbf{1.48 $\pm$ 0.32} & \textbf{13.45 $\pm$ 0.41} & \textbf{11.904  $\pm$ 0.68} & \textbf{45.61 $\pm$ 0.97} & \textbf{118.76 $\pm$ 1.46}
\\ \hline
\multicolumn{1}{|c|}{\multirow{4}{*}{Test RMSE}} & PMF & 0.955  $\pm$ 0.005 & 0.921  $\pm$ 0.008 & 2.163  $\pm$ 0.006 & 0.89  $\pm$ 0.007 & 0.862 $\pm$ 0.005
\\ \cline{2-7} 
\multicolumn{1}{|c|}{} & SVD & 0.922 $\pm$ 0.007 & 0.867 $\pm$ 0.006 & 2.119 $\pm$ 0.004 & 0.795 $\pm$ 0.006 & 0.784 $\pm$ 0.005
\\ \cline{2-7} 
\multicolumn{1}{|c|}{} & BCSVD & 0.924 $\pm$ 0.007 & 0.871  $\pm$ 0.006 & 2.124 $\pm$ 0.006 & 0.798 $\pm$ 0.005 & 0.786 $\pm$ 0.004
\\ \cline{2-7} 
\multicolumn{1}{|c|}{} & BGSVD & \textbf{0.924 $\pm$ 0.005 } & \textbf{0.869 $\pm$ 0.006 } & \textbf{2.121 $\pm$ 0.007} & \textbf{0.795 $\pm$ 0.006 } & \textbf{0.784 $\pm$ 0.005}
\\ \hline
\end{tabular}
\caption{Comparison of basic SGD based PMF with SVD and then with block based SVD that is implemented both on CPU as well as on GPU}
\label{tab:standard}
\end{table*}

\noindent \textbf{Block based GPU accelerated SVD: } In order to fully leverage parallelism and to take advantage of multi-threading capabilities of GPU, we have implemented a GPU variant of Block based SVD (BGSVD). The implementation is specific to CUDA capable hardware supported by Nvidia. We have leveraged 3072 threads which is the maximum supported by the hardware. Accordingly, we have split the MovieLens 1M data set into block matrix of 3072x3072 dimension. For this experimental setup, the entire data matrix is loaded into GPU memory and then logically split into blocks for determining the boundaries of each thread. It is important to note that no memory optimization techniques were used for GPU acceleration. We have not leveraged the shared memory, nor used different caching techniques to fine tune the performance. Hence the performance of the model is due to block based approach alone. Figure \ref{fig:bsvdvsbgsvd} demonstrates the convergence of BGSVD. Considering that the hardware architecture of the CPU and the GPU are significantly different, the time taken by CPU variants need not necessarily be taken as the basis for comparison against the GPU variants. Table~\ref{tab:standard} lists the comparative analysis of the time taken per iteration and the test RMSE for BGSVD.

\section{Conclusions and Future work}
In this work we have proposed a block based approach to SVD that can be scaled and can produce better results when applied in the domain of movie recommender systems. 
The SVD based kernel proved to be providing advantage in RMSE convergence, while the block based approach enables scalability. The proposed technique does not put limitation on the size of the data set. The size of blocks are only limited by the available memory on the computation unit, while there is no limitation on the total number of blocks. The approach provides computational advantage, with respect to time and memory.  By increasing the number of blocks, and by running (in parallel) the exact number of blocks as the number of available cores, it is possible to factorize data sets of any scale. The block based approach can be adapted to run block level factorization on multiple GPUs as well as on distributed systems.

SVD is also useful for incremental data where all the data is not available initially and the new data may arrive after the model building phase. In such a scenario, the models can be incrementally computed with the idea that a preliminary model is computed and then the projection method is used to build incrementally upon that. This method was used in \cite{Sarwar02incrementalsingular} to handle dynamic data sets. It was shown that a projection of additional data can potentially provide a fairly good approximation of the model. A theoretical basis for incremental computation of SVD is provided in \cite{Jengnan}. In \cite{Ross2008}, the incremental SVD is used for visual tracking. However, these models are based on matrix algebra which is not applicable for sparse matrices. With sparse matrices, SVD results in complex numbers that can not be applied for predictions. The block based approach to SVD could be enhanced further to accommodate incremental data. A hierarchical factorization model can be developed by considering the initial data as a block and thereafter the incremental data as new blocks that are independently factorized.  The resultant latent factors can be converged together to arrive at a new model.

The SVD kernel could be further enhanced to factor in temporal changes in user behaviour and time based change in popularity of items. Temporal factors were implemented into the MF algorithm in~\cite{Koren091the} which eventually won the Netflix prize.
Such an implementation can be extended in the context of BMF to take advantage of computational gain and memory optimization. The GPU implementation of the block based SVD can be enhanced further for  memory optimization and data transfer between computational units as proposed in \cite{Tan2018MatrixFO}. The approach can be applied to various domains like text mining~\cite{10.1007/978-3-642-31178-9_15}.\\
\vspace{-2mm}
\bibliographystyle{plain}
\bibliography{bibitems}

\end{document}